\documentclass[runningheads]{llncs}
\usepackage{graphicx,amsfonts,amsmath,color}

\usepackage{algorithm}
\usepackage{algpseudocode}

\newcommand{\RR}{\mathbb{R} }
\def\ncal{{\mathcal N}}

\begin{document}
	\title{Algorithms that get old : the case of generative deep neural networks
	}
	\titlerunning{Generative algorithms that get old}
	%
	\author{Gabriel Turinici
		\orcidID{0000-0003-2713-006X}}

	\authorrunning{G. Turinici}
	\institute{CEREMADE, \\  Universit\'e Paris Dauphine - PSL, CNRS, Paris, France \\
		\email{gabriel.turinici@dauphine.fr}\\
		\url{https://turinici.com} }

	%
	\maketitle              
	\begin{abstract}
Generative deep neural networks used in machine learning, like the Variational Auto-Encoders (VAE), and Generative Adversarial Networks (GANs) produce new objects 
each time when asked to do so with the constraint that the new objects remain similar to some list of examples given as input. However, this behavior is unlike that of human artists that change their style as time goes by and seldom return to the style of the initial creations.
		
We investigate a situation where VAEs are used to sample from a probability measure described by some empirical dataset.
		
Based on recent works on Radon-Sobolev statistical distances, we propose a numerical paradigm, 	to be used in conjunction with a generative algorithm, that satisfies the two following requirements: the objects created do 
not repeat and evolve to fill the entire target probability distribution.
		
	\end{abstract}
	\section{Motivation}

	Consider a distribution $\mu$
	and  $\mu_e = \frac{1}{M}\sum_{\ell=1}^M \delta_{x_\ell}$ an empirical measure sampling this distribution given by a collection of objects 
	$x_m$, $m=1$,..., $M$  where $x_m \in \mathbb{R}^N$ are 
	independent and follow the law $\mu$~; in some sense to be defined latter (cf. discussion on statistical distances) $\mu_e$ is close to $\mu$;
	we focus on generative deep neural network architectures that, given $\mu_e$ can produce samples from the distribution $\mu$.
	One such neural network class are the 
	Variational Auto Encoders (cf. \cite{kingma_introduction_2019,CWAE} for an introduction) that, after some training, output two functions (that in practice are implemented as neural networks): the encoder function $E: x \in \mathbb{R}^N\mapsto z \in \mathbb{R}^L$ and the decoder function $D: z \in \mathbb{R}^L \mapsto y \in \mathbb{R}^N$; the decoder function has the property that the image of a multi-dimensional Gaussian on the latent space $\mathbb{R}^L$ through $E$ is close to $\mu_e$ thus to $\mu$.	
	Some recent proposals to construct such a VAE are 
	presented in \cite{turinici_radonsobolev_2021}, which will also be our inspiration for the statistical distance used in this work (see also \cite{szekely_energy_2013}).	
The quality of a VAE is given 
\begin{enumerate}
	\item by the proximity of the $D\circ E$ to the identity operator (at least on the support of the target measure $\mu$);
	\item and the small distance between target distribution $\mu$ and $D(\ncal(O,I_L))$ (here $\ncal$ is the $L$-dimensional standard Gaussian);	
\end{enumerate}
However, although in general the VAEs (same thing applies to the Generative Adversarial Networks - GANs ) obtain very good quality results by the previous criteria, the sampling  performed at the exploitation phase is, because of the construction, done in an independent way: each time a new $y \sim \mu$ is required, a $z \simeq \ncal_L$ is sampled and $D(z)$ computed. But such a procedure is at odds with what we observe in real life: the painters do not paint the same landscape again (but still paint pictures), the musical composers' productions vary in style over the years, etc, in general some evolution is witnessed with time. Such a phenomena is probably due to  {\bf taking into account the objects previously created}. Our goal is to be able to mimic such an evolution and propose a generative algorithm that
\begin{enumerate}
	\item is able to create new objects from some target distribution $\mu$ (that for VAEs and GANs is the latent distribution);
	\item is able to "recall" having created previous objects; 
{\bf This second point will therefore synthetically induce an "artificial age" for an AI because the process is irreversible.} 
\end{enumerate}

A non-aging generative algorithm, when asked to produce, e.g. one new result will likely produce the same object (or similar) over and over again: think of the situation of a standard $1D$-Gaussian: most likely the origin will be drawn over and over again, one has to wait a long time to obtain let's say, a value at $3$ standard deviations from the mean.
The main goal of this paper is to speed up this waiting time.
The advantages of such a process is to allow some "maturation" for the results i.e. to be able to create new results, not the same ones again; this comes at the price of a irreversibility and additional computation cost.

\subsection{Relation to previous literature}
Technically, our proposal has some similarities with different areas in computational statistics: first one can invoke the "vector quantization" procedures (see \cite{kohonen_learning_1995,r_gray_vector_1984,book_quantization_measures} and references therein) that, given a distribution, find a set of objects that represent it as a sum of Dirac masses. However, there the technical solution (Voronoi diagrams for instance) 
is naturally oriented to use for probability measures (or more generally finite positive measures) which is not our situation (our effort involves signed measures); in the same vein see also \cite{measure_quantization} in the context of machine learning algorithms. On the other hand some efforts have been made to generalize the quantile idea to multi-dimensional distributions; in a one-dimensional situation our procedure and these techniques give similar results but they diverge as soon as the dimension is increased, see \cite{glazer_q-ocsvm_2013,FRAIMAN20121,multivariate_quantiles}.

On another hand and from a completely different perspective, the notion of "age" of a task in a queue is used in scheduling to ensure execution of low priority processes, see \cite{silberschatz_operating_2018}.

\subsection{Technical goal of the paper}

Continuing the works above, and given the discussion on the generative algorithms, we need a procedure that can incrementally find a good representation of a target measure $\mu$ as a number of $K$ Dirac masses ($K$ is given and fixed) centered at 
some $x_k$, $k=1,...,K$
while taking into account a set of points $Y = (y_j)_{j=1}^{K_p}$ already available. 
The points $Y$ are called {\bf historical points}. 
To put it otherwise we want to find the multi-point $X=(x_k)_{k=1}^K \in \RR^{N \times K}$ ($k=1,...,K$) that minimizes the distance 
from the total empirical distribution $\frac{\sum_{k=1}^{K_p} \delta_{y_k} + 
	\sum_{l=1}^K \delta_{x_k}}{K_p+K}$ to the target measure $\mu$ (here the points $y_k$ are not submitted to optimization); this can be written as minimizing the distance
$d(\delta_X,\eta)^2$
from the distribution
$\delta_X=\frac{1}{K}\sum_{l=1}^K \delta_{x_k}$ to the signed measure
\begin{equation}
\eta= \frac{(K_p+K) \mu- K_p \delta_Y}{K_P+K}, \text{ where } \delta_Y= \frac{1}{K_p}\sum_{k}^{K_p} \delta_{y_k}. 
\label{eq:targetsignedmeasure}
\end{equation}

We present in section \ref{sec:theory} our choice of distance $d$ and a theoretical result ensuring that, under appropriate hypotheses, the minimum with respect to $X$ exits.
The algorithm to find such a minimum is presented in section \ref{sec:algo}  together with some numerical results.
Final remarks are the object of section \ref{sec:discussion}.

\section{Theoretical results} \label{sec:theory}

In order to present the theoretical framework we need to define the distance $d$ between signed measures $\zeta$ and $\eta$. Note that 
in fact $\zeta$ is a probability measure and the total mass of both is set to $1$.

We will take a kernel-based metric given as follows: choose $h(\cdot)$ a conditionally negative definite kernel (see \cite{AnnalsStat_endist13} for the precise definition and an introduction),  taken here to be $\sqrt{a^2+|x|^2}- a$ for some given constant $a\ge 0$), see \cite{Deshpande_2019_CVPRmax,turinici_radonsobolev_2021} for some use cases in machine learning; for any $\eta_1$, $\eta_2$ signed measures such that $\int (1+|X|) \eta_i(d X)< \infty$ ($i=1,2$) we define~:
\begin{equation}
d(\eta_1,\eta_2) =  \sqrt{\int \int -h(|X-Y|) (\eta_1-\eta_2)(dX) (\eta_1-\eta_2)(dY)}.
\label{eq:defdistance}
\end{equation}

The fact that the quantity inside the square root is positive is a consequence of the fact that $h$ is a conditionally negative definite kernel.

Note that in particular, if both $\eta_1$ and $\eta_2$ are sums of (signed) Dirac masses 
such that $\eta_1-\eta_2 =\sum_{k=1}^{K}  p_k \delta_{z_k}$ (with $K< \infty$) then
equation \eqref{eq:defdistance} can be written (see \cite{turinici_radonsobolev_2021})~:
\begin{equation}
	d(\eta_1,\eta_2)^2 = - \sum_{k,\ell=1}^{K} p_k p_\ell 
	h(|z_k-z_\ell|).	\label{eq:distance_as_sum_diracs}
\end{equation}

Once the distance is defined, a legitimate question is whether, given a target signed measure $\eta$ one can indeed find a uniform sum of Dirac masses $\zeta$ that minimizes 
$d(\zeta,\eta)^2$. This question is settled in the following
\begin{proposition} Suppose $K$ is a fixed positive integer.
Let $\eta$ be a signed measure such that $\int (1+|X|) \eta(d X)< \infty$~.
For any vector $Z=(z_j)_{j=1}^J \in \RR^{N\times J}$ denote 
\begin{equation}
	\delta_Z := \frac{1}{J}\sum_{j=1}^J \delta_{z_j}, \ 
	f(Z):= d \left(\delta_Z ,\eta \right)^2.
\end{equation}
Then the minimization problem~:
\begin{equation}
 \inf_{X=(x_k)_{k=1}^K \in \RR^{N\times K}} f(X) 
\label{eq:minimization}
\end{equation}
admits at least one solution.
\end{proposition}
\begin{remark}
The previous result only states the existence of a solution, the uniqueness is not necessarily true as one can observe by taking e.g. a rotation invariant measure: any solid rotation of a minimum will still be a minimum.
\end{remark}
\begin{proof}
Let us denote
\begin{equation}
m_\eta:= \inf_{X=(x_k)_{k=1}^K \in \RR^{N\times K} } f(X).
\end{equation}

Take a point $X$ such that $f(X)\le m_\eta + 1$ (the existence of $X$ is guaranteed by the definition of $m_\eta$). Then (denoting by $0$ the null vector in 
$\RR^{N \times K}$):
\begin{equation}
 m_\eta + 1 \ge f(X) =  d \left(\delta_X ,\eta \right)^2 \ge 
 \frac{d \left(\delta_X ,\delta_0 \right)^2 -2 d \left(\delta_0 ,\eta \right)^2 }{2}, 
\end{equation}
which implies 
\begin{equation}
d \left(\delta_X ,\delta_0 \right)^2 \le 
2(	m_\eta + 1) + 2 d \left(\delta_0 ,\eta \right)^2.
\end{equation}
But, using equation \eqref{eq:distance_as_sum_diracs}, we obtain~: 
\begin{eqnarray}
& \ &	d \left(\delta_X ,\delta_0 \right)^2 
= 
 \frac{2}{K}\sum_{k=1}^K h(|x_k|)
- \frac{1}{K^2}\sum_{k,k'=1}^K h(|x_k-x_{k'}|)
\label{eq:ineqh1}
\\ & \ &
\ge  \frac{2}{K}\sum_{k=1}^K h(|x_k|)
- \frac{1}{K^2}\sum_{k,k'=1, k \neq k'}^K \left[a+ h(|x_k|)+h(|x_{k'}|)\right]  
\\ & \ &
\ge  \frac{2}{K^2}\sum_{k=1}^K h(|x_k|)-a,
\label{eq:ineqh2}
\end{eqnarray}
where for the passage from \eqref{eq:ineqh1} to \eqref{eq:ineqh2} we used the inequality
$ h(|x-y|) \le h(|x|) + h(|y|) + a$ true for any $x,y \in \RR^N$.  
We obtain that, when $f(X)\le m_\eta + 1$ there exists a constant $C_0 = 
K^2 \left( a/2+ m_\eta + 1 + d \left(\delta_0 ,\eta \right)^2 \right)$
 such that
$\sum_{k=1}^K h(|x_k|) \le C_0$;
therefore any minimizing sequence $(X_n)_{n\ge 1}$ (that is, any sequence such that 
$\lim_{n\to \infty}f(X_n)= m$) is bounded. This sequence will have a sub-sequence $(X_{n_k})_{k\ge 1}$ that is convergent to some $X^\star$; but since the distance is continuous, we obtain that $f(X^\star)=m$ which means that $\delta_{X^\star}$ is a solution of the minimization problem \eqref{eq:minimization}. \qed
\end{proof}

\section{Algorithm and numerical results} \label{sec:algo}

\subsection{Algorithm formulation}

Consider now a target distribution $\mu$ and a set of previously constructed points $Y$ (of cardinal $K_p$);  these {\bf historical points} are explicitly known; we will propose  an algorithm that, given a number $K$ of points to be constructed, will find a multi-point $X\in \RR^{N \times K}$ such that the overall measure
$\delta_{X \cup Y}$ minimizes the distance to the target measure $\mu$. 

\begin{algorithm}
	\caption{\!\! History aware (signed measure) compression algorithm : HAW-C}
	\label{alg:has}
	\begin{algorithmic}[1]
		\Procedure{HAW-C}{}
		\State $\bullet$ set {batch size $B$, parameter $a=10^{-6}$,
		\State $\bullet$ load the historical points $y_k$, $k=1,...,K_p$}
		\State $\bullet$ initialize points $x_k$, $k=1,...,K$ sampled at random from $\mu$, denote $X=(x_k)_{k=1}^K$ (considered as vector in $\RR^{N \times K}$) 
		\While{(max iteration not reached)}
		\State $\bullet$ sample $z_1,...,z_B \sim \mu$ (i.i.d).
		\State $\bullet$ compute the global loss~\footnote{The global loss is the distance from 
		$\delta_X=\frac{1}{K}\sum_{l=1}^K \delta_{x_k}$ to the signed measure
		$\eta$.} using  formula \eqref{eq:distance_as_sum_diracs}~: 
	
$L(X) := d\left( \frac{1}{K}\sum_{l=1}^K \delta_{x_k}, 
\frac{K_p+1}{B}\sum_{b=1}^B \delta_{z_b} - \sum_{j=1}^{K_p} \delta_{y_j}
 \right)^2$;

		\State $\bullet$ backpropagate the loss $L(X)$ in order to minimize $L(X)$ and update $X$.		 
		\EndWhile\label{euclidendwhile}
		\EndProcedure
	\end{algorithmic}
\end{algorithm}

\subsection{Numerical results}
	
A Python code implementing the algorithm in both history unaware and history aware compression modes can be consulted at \cite{code_compression_LOD22}.

\subsubsection{History unaware compression of a 2D Gaussian mix distribution}

\ 
\newline

\begin{figure}
	\includegraphics[width=0.5\textwidth]{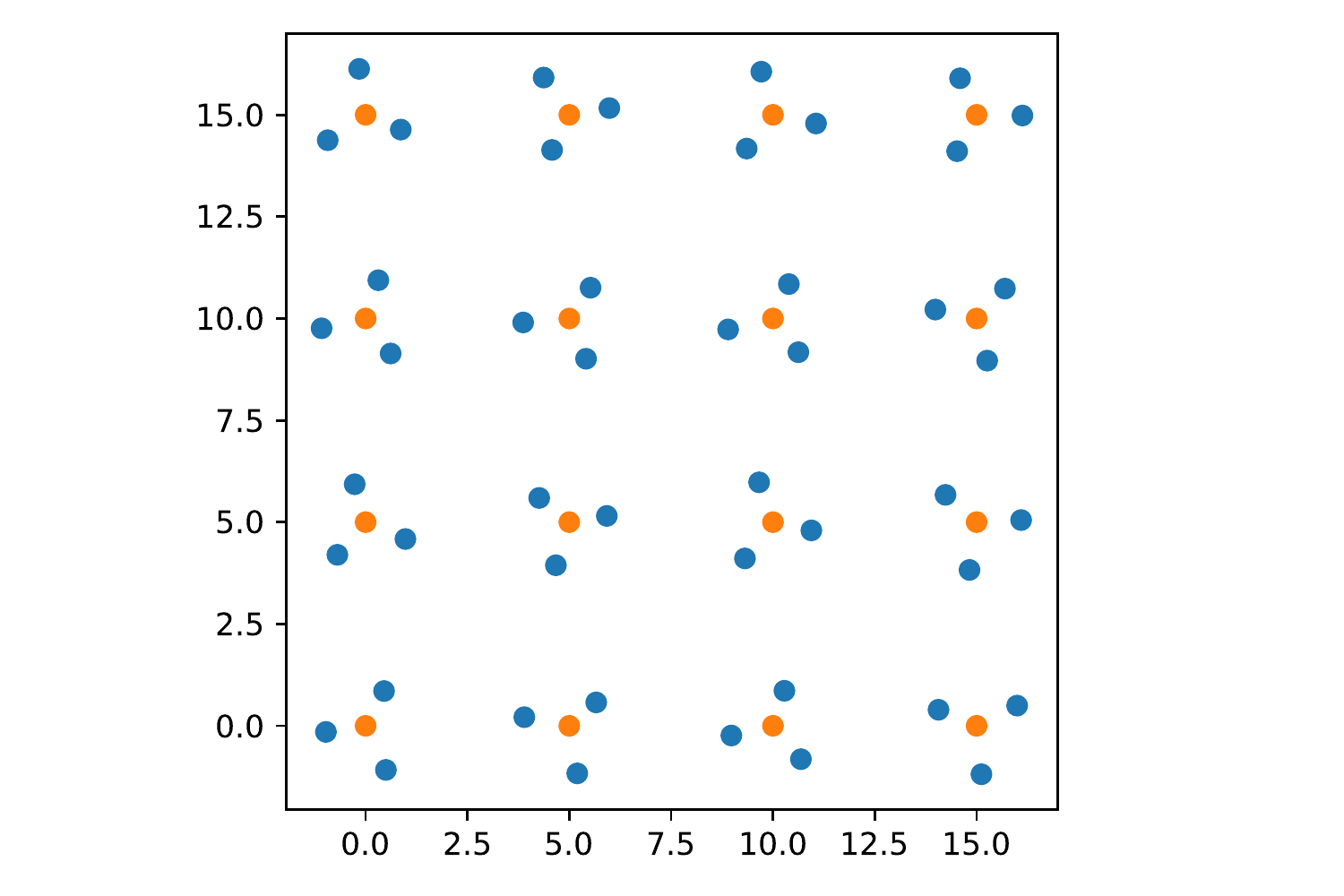}
	\includegraphics[width=0.5\textwidth]{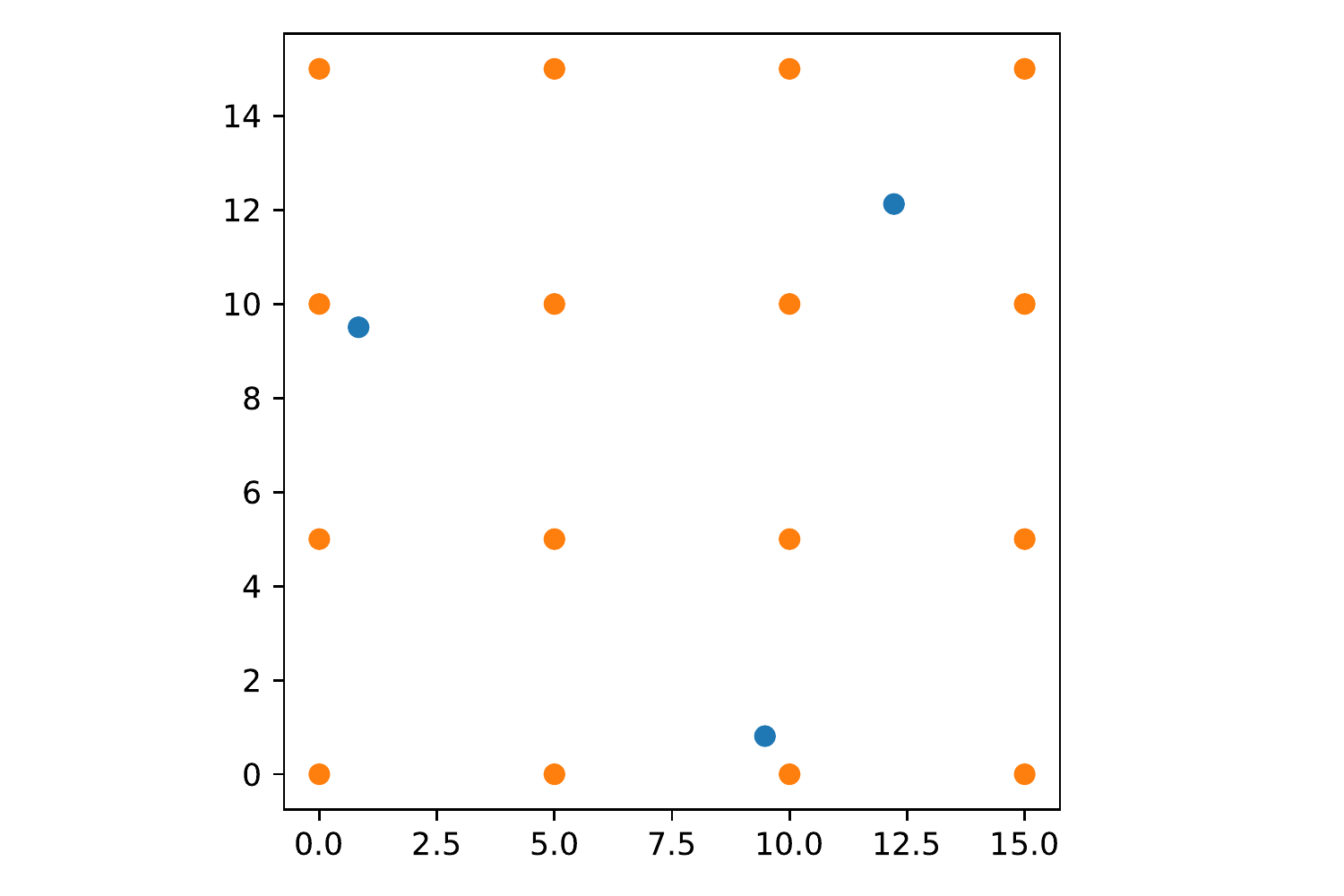}
	\caption{Test without any historical points, $K_p=0$. 
		An example of compression for an uniform Gaussian mixture 
		of $16$ Gaussians centered on points of a $4\times 4$ grid (red points are the centers of the Gaussians, blue points are the compressed points). We used $K$ 	points to summarize the distribution~: $K=48$ 
		({\bf left image}) or  $K=3$ ({\bf right image}).
Good quality results are obtained as the algorithm "understands" the mixing structure: for instance for $K=48$ the algorithm allocates precisely $3$ points per Gaussian mixture term.	
} \label{fig:odegaussian}
\end{figure}

We first test the algorithm without any historical points i.e., $K_p=0$. 
When the target measure is positive, the HAW-C algorithm \ref{alg:has} allows to compress any given (probability) measure, as illustrated in figure \ref{fig:odegaussian} for the situation of a uniform mixture of $16$ lattice-centered $2D$ normal variables. Good results are obtained: without any previous knowledge, the algorithm can unveil the mixing structure and allow a coherent compression.

\subsubsection{History aware multi-dimensional Gaussian compression and application to generative algorithms}

\ \newline

\begin{figure}
\begin{center}
	\includegraphics[width=0.5\textwidth]{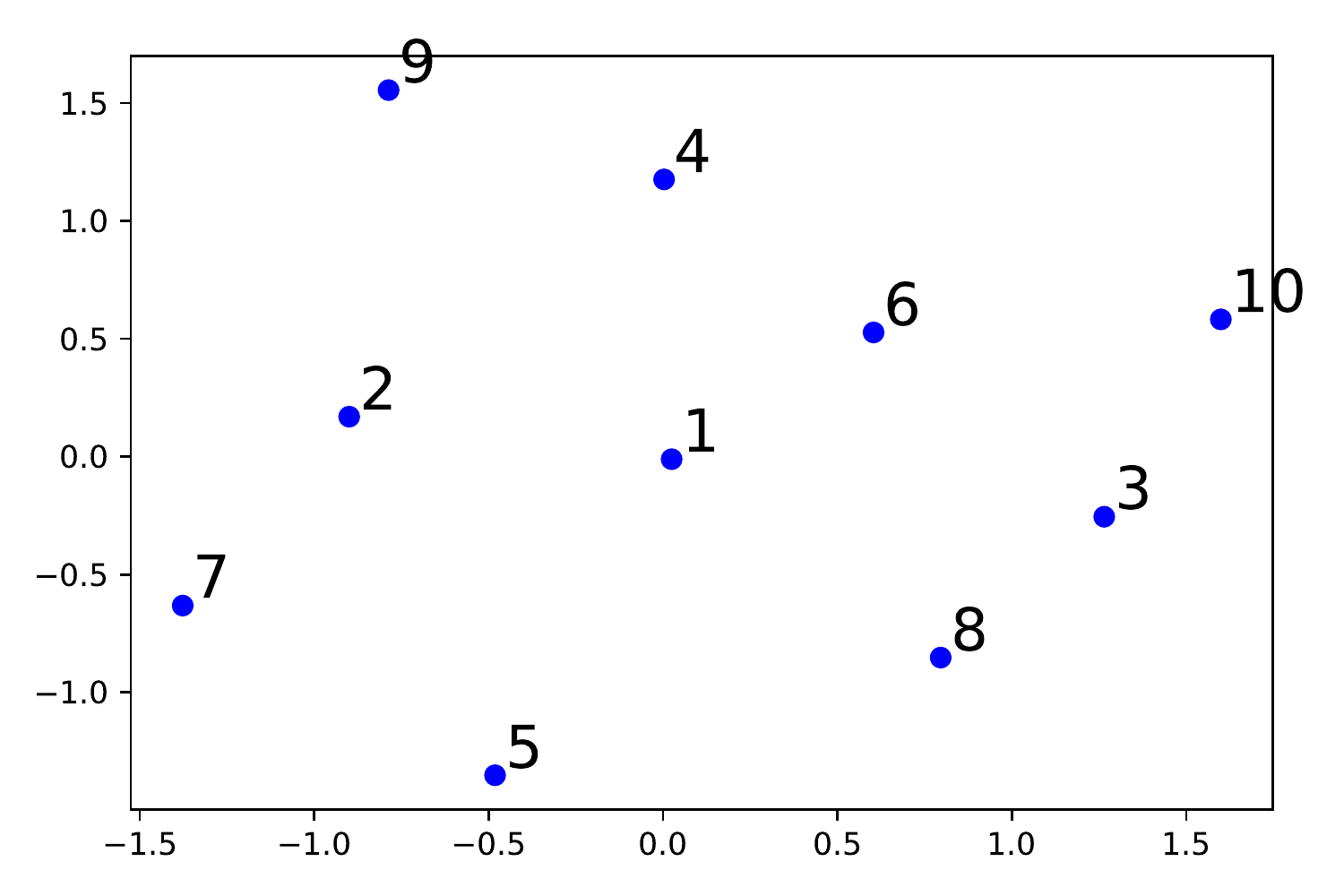}
\caption{An example of recursive compression of  a $2D$ standard Gaussian (see text for details). {\bf Left image:} the result of the compression after $10$ iterations. Each point $u_i$ is labeled by its corresponding index $i$ when it was found.
} \label{fig:1Dgaussian_signed}
\end{center}
\end{figure}

We move now to a test where incremental compression is performed: we consider a $2D$ Gaussian centered at origin. First we compress it with a single point $u_1$; then we use $K_p=1$ and $y_1=u_1$ as history and compress the signed measure : initial Gaussian minus the first obtained point $u_1$, as detailed in equation \eqref{eq:targetsignedmeasure}, with another supplementary point $u_2$; then consider $K_p=2$ and $y_i=u_i$ ($i=1,2$) and compress the Gaussian measure minus the sum of Dirac masses in $u_i$ with another point $u_3$; the procedure is then continued recursively, each step being an application of the algorithm \ref{alg:has}. The results are presented in figure \ref{fig:1Dgaussian_signed}.

\begin{figure}
	\begin{center}
		\includegraphics[width=0.45\textwidth]{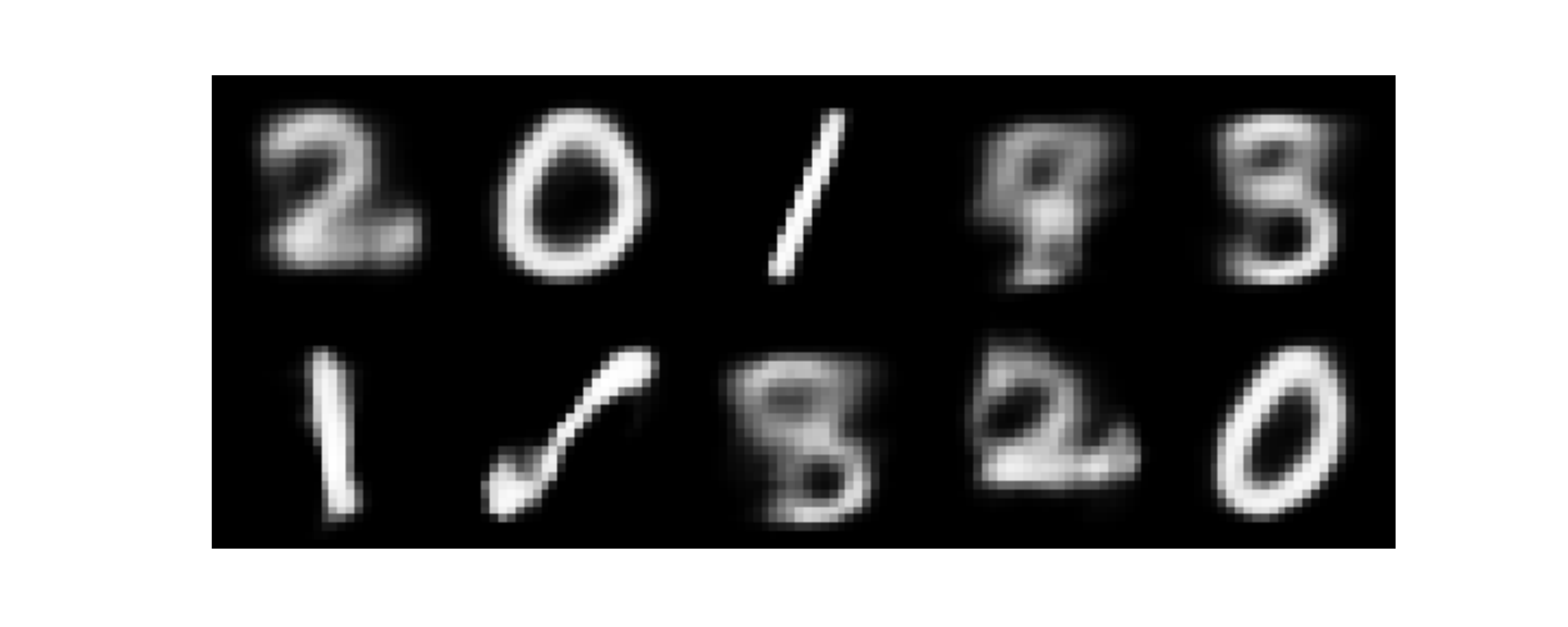}
		\includegraphics[width=0.45\textwidth]{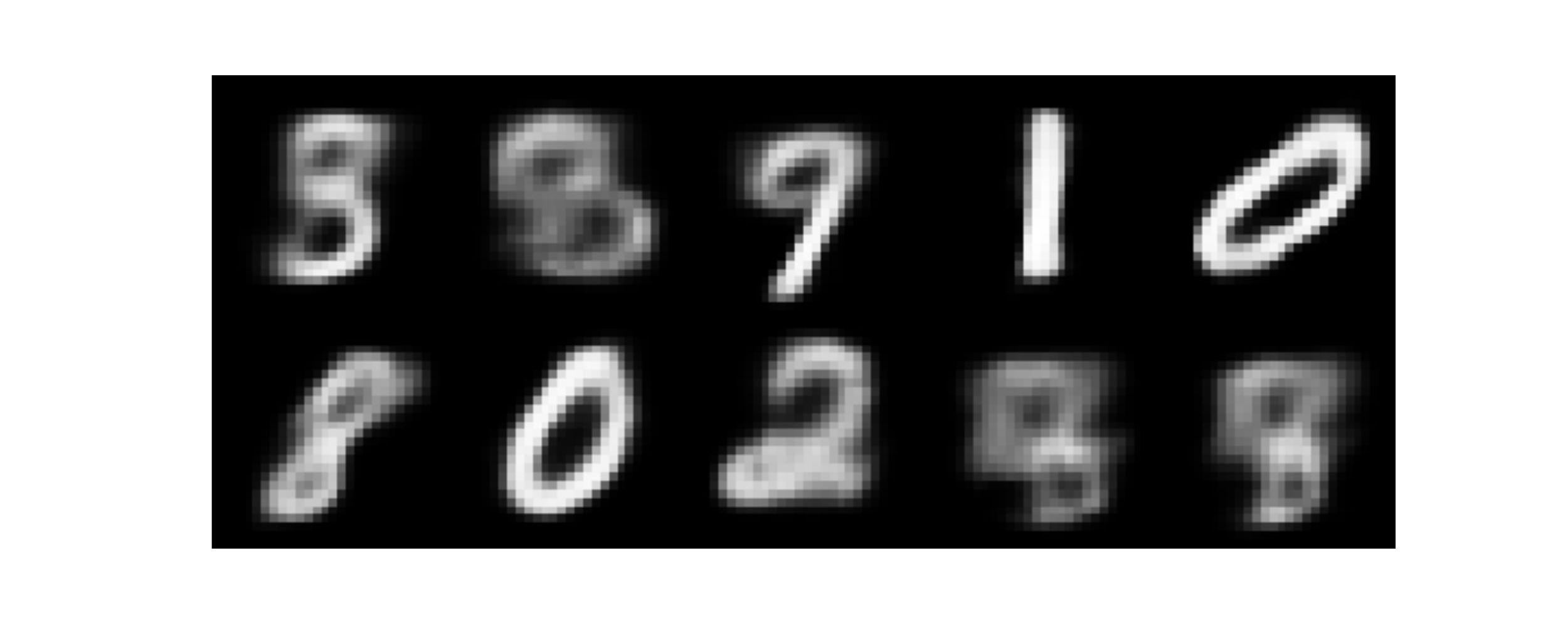}
		\caption{Images from \cite{cvae_tensorflow_jan22} obtained by taking either a random sampling of $10$ points from a 2D Gaussian (left image) or the sampling obtained in figure \ref{fig:1Dgaussian_signed} (right image). The decoder is the one obtained by a run of the code from \cite{cvae_tensorflow_jan22}. The right image appears more faithful of the database. Improved quality results are available in figure \ref{fig:vae_sampling_results512}.}. \label{fig:vae_sampling_results}
	\end{center}
\end{figure}

In order to test these results on a practical case, we used the CVAE procedure from the Tensorflow documentation \cite{cvae_tensorflow_jan22} with default parameters (except that we used $100$ epochs instead of $10$ because the results with $10$ epochs are very fuzzy). The code was executed once in order to construct the encoder/ decoder networks and then the sampling was done in the latent space using either a multi-dimensional sampling of $10$ objects or an incremental sampling;
once the sampling is done the data is propagated through the decoder network and the resulting 
images are presented in figure \ref{fig:vae_sampling_results}. We note that the history aware sampling retains a good diversity with respect to the uniform sampling and avoids some repetitions:

- the figures $1$ and $2$ that are repeated in the propagated random samples and only appear once in the incremental sampling; 

- the figure $9$ appears only very unclearly in the left sample and more clearly in the right sample

- the incremental sampling avoids the symbol in row $2$ column $2$ (left image) which is not a figure.

\begin{figure}
	\begin{center}
		\includegraphics[width=0.45\textwidth]{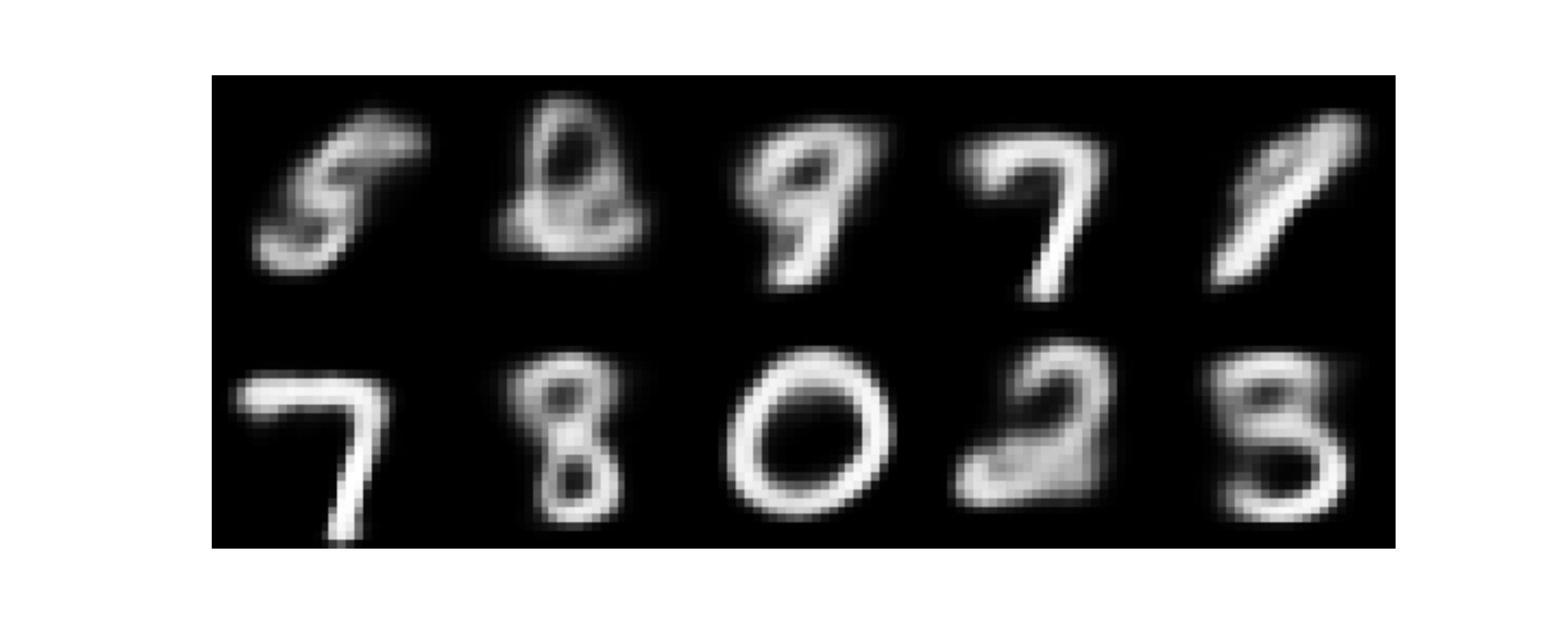}
		\includegraphics[width=0.45\textwidth]{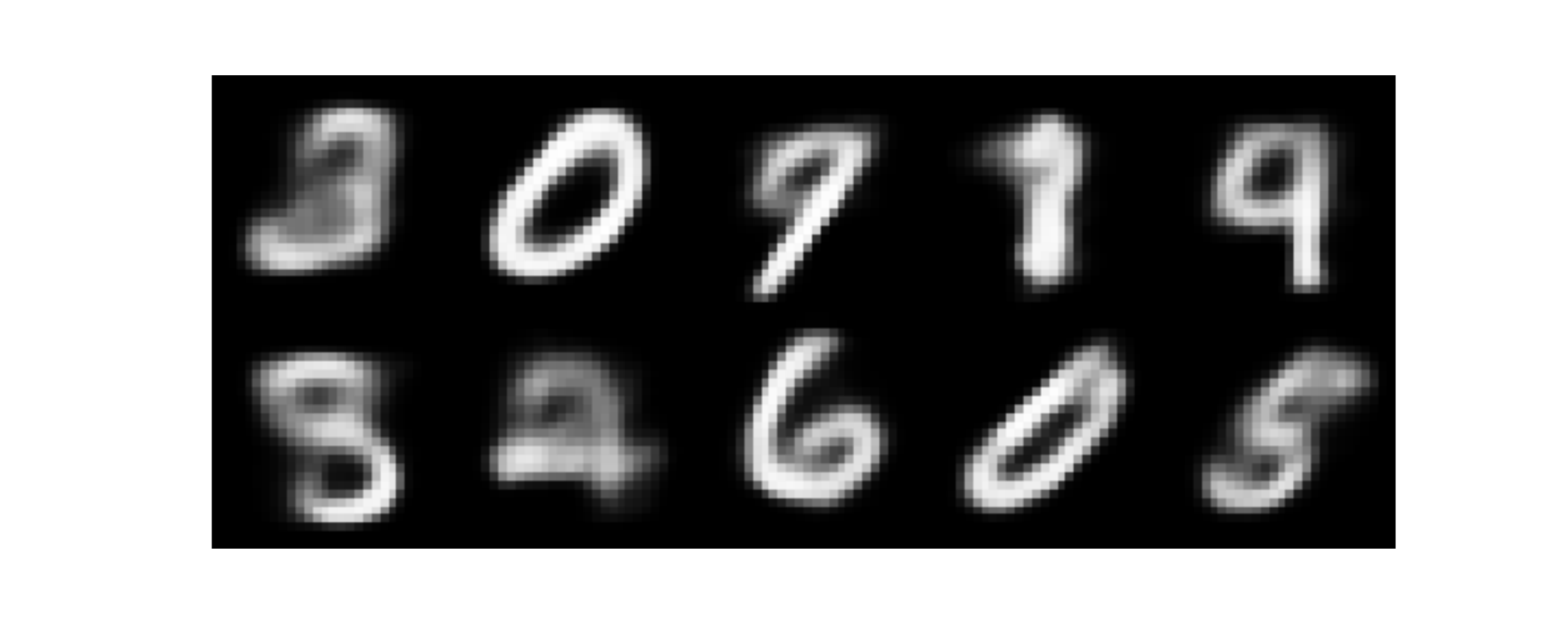}
		\caption{Results of the same procedure as in figure \ref{fig:vae_sampling_results}
			but obtained for an improved network ($512$ filters everywhere and $20$ epochs):
			{\bf left image~:}  random sampling; {\bf right image~:} decoding of the incremental sampling from figure \ref{fig:1Dgaussian_signed}.  The left figure has several repetitions (for instance figure $7$) that are absent from the right figure but more importantly, some figures abondant in the database and not present in the left figure appear in the right one, like the figures $1$ and $6$.}. \label{fig:vae_sampling_results512}
	\end{center}
\end{figure}

Note that this out-of-the-box C-VAE is not good enough to
make figures too precise which explains large numbers of fuzzy images - resembling to a $8$, $6$ or $9$- present in both results). To improve the result, we re-run the CVAE for $20$ epochs but increased, as recommenced in the documentation, all 'filters' numbers to $512$ (instead of $32$ or $64$ in the initial setup). We obtain the results in figure \ref{fig:vae_sampling_results512}: the quality of the VAE is indeed increased and the same conclusions hold for the comparison of the i.i.d. sampling with the incremental sampling.

\section{Final remarks} \label{sec:discussion}

We explored in this work the construction of objects from generative algorithms (like VAEs and GANs); more specifically the construction was incremental in the sense that each new sampling from the latent space considers the previous samples, called the {\bf history}  and tries to both respect the desired target distribution of the latent space but also stays away from the points already sampled. We described some theoretical properties of the procedure and tested it both on general data and on a C-VAE benchmark.

More experiments are required to characterize fully the applicability domain of  the proposed procedure, but the present results provide encouraging arguments to do so.

\bibliographystyle{splncs04}
\bibliography{genbib}
	
\end{document}